\documentclass[conference]{IEEEtran}
\IEEEoverridecommandlockouts
\usepackage{cite}
\usepackage{amsmath,amssymb,amsfonts}
\usepackage{algorithmic}
\usepackage{graphicx}
\usepackage{textcomp}
\usepackage{xcolor}
\usepackage{tabularx}
\usepackage{multirow}
\usepackage{subfigure}

\newcolumntype{M}[1]{>{\centering\arraybackslash}m{#1}}

\def\BibTeX{{\rm B\kern-.05em{\sc i\kern-.025em b}\kern-.08em
    T\kern-.1667em\lower.7ex\hbox{E}\kern-.125emX}}
\begin{document}

\title{Enhanced Knowledge Injection for \\Radiology Report Generation\\}
\author{
\IEEEauthorblockN{Qingqiu Li$^{1}$, Jilan Xu$^{1}$, Runtian Yuan$^{1}$, Mohan Chen$^{1}$, Yuejie Zhang$^{1,*}$, Rui Feng$^{1,*}$, Xiaobo Zhang$^{2}$, Shang Gao$^{3}$}

\IEEEauthorblockA{
$^{1}$ \textit{Fudan University, Shanghai, China}}
\IEEEauthorblockA{
$^{2}$ \textit{Children’s Hospital of Fudan University, Shanghai, China}}
\IEEEauthorblockA{
$^{3}$ \textit{Deakin University, Victoria, Australia}}
}

\maketitle

\begin{abstract}
Automatic generation of radiology reports holds crucial clinical value, as it can alleviate substantial workload on radiologists and remind less experienced ones of potential anomalies. Despite the remarkable performance of various image captioning methods in the natural image field, generating accurate reports for medical images still faces challenges, i.e., disparities in visual and textual data, and lack of accurate domain knowledge. To address these issues, we propose an enhanced knowledge injection framework, which utilizes two branches to extract different types of knowledge. The Weighted Concept Knowledge (WCK) branch is responsible for introducing clinical medical concepts weighted by TF-IDF scores. The Multimodal Retrieval Knowledge (MRK) branch extracts triplets from similar reports, emphasizing crucial clinical information related to entity positions and existence. By integrating this finer-grained and well-structured knowledge with the current image, we are able to leverage the multi-source knowledge gain to ultimately facilitate more accurate report generation. Extensive experiments have been conducted on two public benchmarks, demonstrating that our method achieves superior performance over other state-of-the-art methods. Ablation studies further validate the effectiveness of two extracted knowledge sources.
\end{abstract}

\begin{IEEEkeywords}
Radiology Report Generation, Knowledge Injection, Weighted Concept Knowledge, Multimodal Retrieval Knowledge
\end{IEEEkeywords}

\section{Introduction}

Medical imaging has been widely employed in clinical practice to assist in disease diagnosis. However, the extensive interpretation of radiological images and the composition of diagnostic reports impose a significant burden on radiologists. Furthermore, less experienced radiologists might overlook certain anomalies in radiological images. To alleviate the substantial workload on radiologists and remind those with less experience of potential anomalies, the automatic generation of radiology reports holds crucial clinical value.
\begin{figure}[t]
\centering
\includegraphics[width=0.48\textwidth]{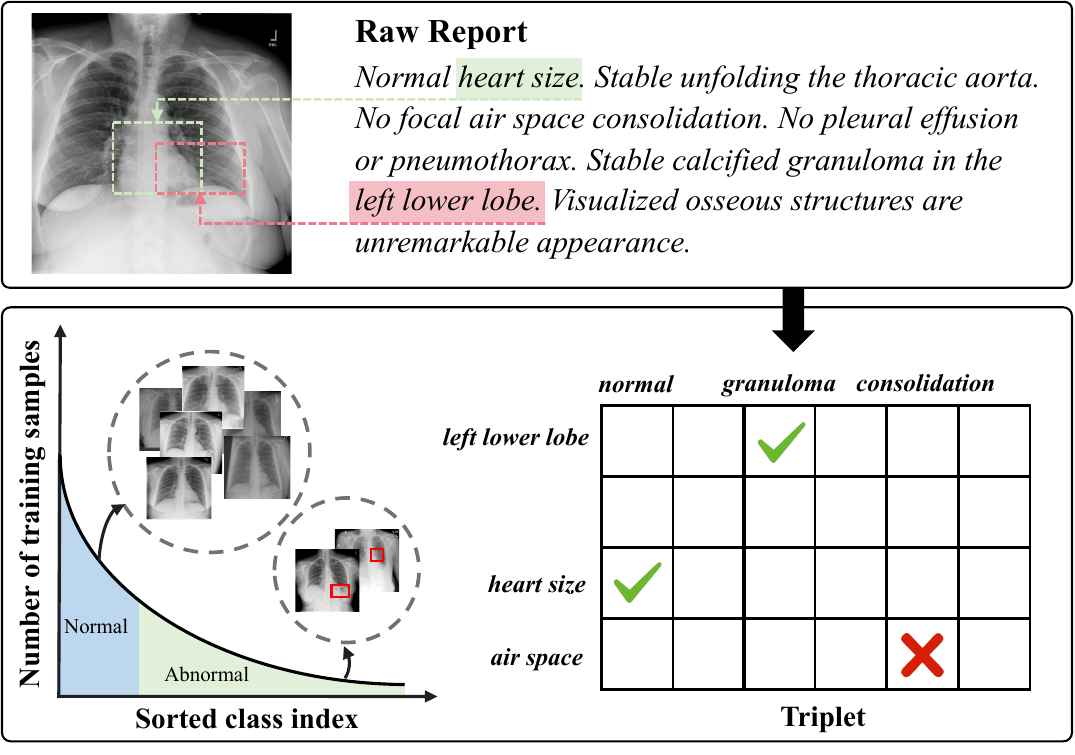}
\caption{An illustration of imbalanced positive and negative samples in medical data, and positional and existential information in radiology report.}
\label{fig:illustration_samples}
\vspace{-0.3cm}
\end{figure}

In recent years, automatic radiology report generation has attracted extensive research interest. The majority of existing methods~\cite{wang2018tienet,jing2018automatic,li2018hybrid} follow a paradigm of conventional image captioning approaches, employing an encoder-decoder framework. These methods typically utilize Convolutional Neural Networks (CNNs) as encoders and Recurrent Neural Networks (RNNs) as decoders. More recently, Transformer-based architectures~\cite{liu2021exploring,chen2021cross,qin2022reinforced} have also been widely adopted. The outcomes of previous endeavors have demonstrated that the encoder-decoder paradigm yields commendable performance gains in the field of radiology report generation.

Despite these advancements, there remain several salient issues that cannot be ignored. (1) Disparities in visual and textual data. In terms of visual data disparities, as shown in Fig.~\ref{fig:illustration_samples}, there are often more normal images than abnormal ones in datasets, leading to an imbalanced visual distribution. This imbalance weakens the model's ability to accurately pinpoint anomalous regions. Regarding textual data disparities, when writing medical reports, radiologists tend to describe all elements within images, resulting in an over-representation of descriptions for normal regions throughout the entire report. This textual imbalance hinders the model's capability to describe specific critical abnormalities. (2) Lack of domain knowledge. Unlike natural image captioning, radiology report generation necessitates a high level of domain expertise. However, relying solely on encoder-decoder structures fails to precisely integrate domain expertise into report generation, making it challenging to be widely adopted as a clinical decision support. Current knowledge injection approaches~\cite{yang2022knowledge,li2023dynamic} introduce a substantial amount of unrelated or even contradictory knowledge into the current image. These approaches tend to emphasize only entity knowledge while ignoring the positional information which is crucial in clinical practice.

To address the aforementioned issues, we propose an enhanced knowledge injection framework for radiology report generation. Our framework is composed of two branches: a Weighted Concept Knowledge (WCK) branch and a Multimodal Retrieval Knowledge (MRK) branch. These branches are designed to incorporate more accurate and noise-suppressing domain knowledge. The combination of such two types of knowledge with the current image features is achieved through the Mixture of Knowledge (MoK) module. The fused knowledge-enhanced feature serves as the input to the Transformer for the final report generation. 
Specifically, the WCK introduces weighted clinical medical concepts, mitigating the problem of extreme data sample imbalance while incorporating domain knowledge. In contrast, the MRK transforms retrieved reports into triplets, introducing finer-grained and well-structured knowledge. 

Our primary contributions are outlined as follows:
\begin{itemize}

\item We present an enhanced knowledge injection framework for radiology report generation, addressing the challenge of imbalanced samples while incorporating well-structured and low-noise knowledge.

\item We propose two distinct branches to extract different types of knowledge. The WCK branch is responsible for incorporating weighted clinical concepts. And the MRK branch extracts triplets with a novel format from similar reports, emphasizing vital clinical information about entity positions and presence.

\item Our proposed knowledge injection framework achieves superior performance over other state-of-the-art methods. The quantitative and qualitative results demonstrate the improved accuracy and fluency of reports attributed to the two knowledge sources.

\end{itemize}

\section{Related Work}
A number of deep learning networks have been proposed for  automatic radiology report generation. TieNet~\cite{wang2018tienet} employed multi-layer attention to highlight meaningful words and image regions. CoAttn~\cite{jing2018automatic} incorporated disease labels into the generation process and utilized hierarchical LSTM to produce disease-related reports. HRGR~\cite{li2018hybrid} introduced a retrieval-generation approach, deciding whether to generate a new sentence or retrieve existing templates for different topics. 
R2GenCMN~\cite{chen2021cross} and CMM+RL~\cite{qin2022reinforced} employed trainable cross-modal feature storage matrices as intermediaries between visual and textual patterns to enhance global modality alignment. Furthermore, to generate more coherent and consistent reports, reinforcement learning~\cite{li2018hybrid,qin2022reinforced} was used to directly optimize required metrics.

\begin{figure*}[t]
\centering
\includegraphics[width=1\textwidth]{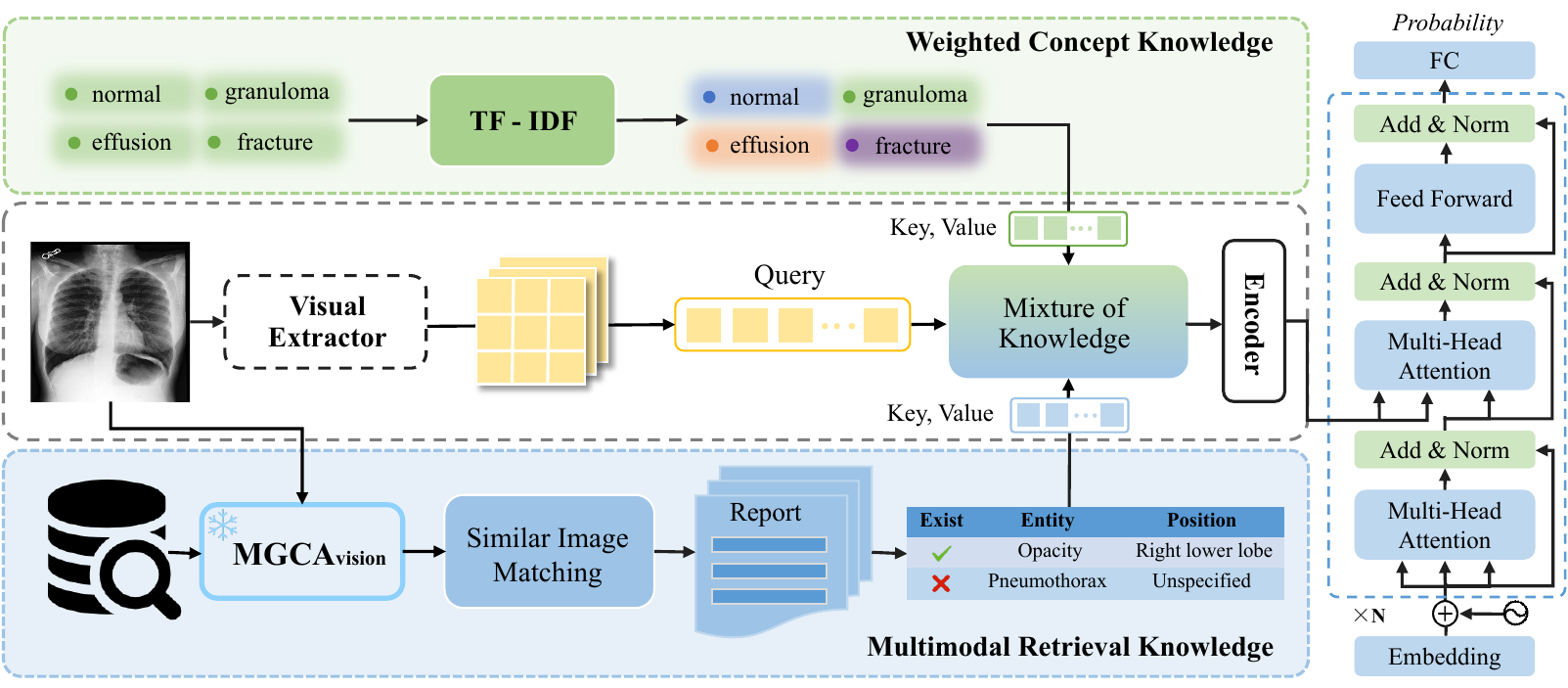}
\caption{Overall architecture of our proposed framework. The knowledge extracted from the Weighted Concept Knowledge branch and the Multimodal Retrieval Knowledge branch is fused with the current image feature, and then used as input for the Transformer to generate the radiology report.}
\label{fig:framework}
\end{figure*}

Simultaneously, some efforts focus on assisting report generation with additional knowledge. To enhance the report generation model's ability to understand medical domain knowledge, KERP~\cite{li2019knowledge} and MEET~\cite{zhang2020radiology} constructed these abnormal images based on prior medical knowledge, modeled intrinsic relationships, and generated radiology reports from abnormal images. PPKED~\cite{liu2021exploring} simulated the radiologists' workflow by exploring and refining both posterior (visual information) and prior knowledge (medical graph and retrieved reports). Know Mat~\cite{yang2022knowledge} introduced a framework based on general and specific knowledge. General knowledge was derived from a pre-constructed knowledge graph, while specific knowledge was obtained from retrieved similar reports. 
However, the aforementioned methods treat all knowledge as equally significant during knowledge injection process, which is sub-optimal for radiology report generation tasks that particularly focus on abnormalities. Moreover, these methods fail to emphasize crucial clinical information about the location and presence of entities, leading to the introduction of numerous irrelevant or even contradictory noisy knowledge. Different from these existing works, we propose an enhanced knowledge injection framework to introduce fine-grained and low-noise knowledge.

\section{Methodology}

\subsection{Overview}

The overall architecture of our framework is illustrated in Fig.~\ref{fig:framework}. It adopts an encoder-decoder structure, composed of a Weighted Concept Knowledge branch and a Multimodal Retrieval Knowledge branch. Given a radiograph $I$, the image feature $F_I$ is extracted by visual extractor.
The objective is to generate a descriptive radiology report $ R=\{ y_1, y_2, \ldots ,y_{N_R} \}$, where $y_i$ is a word token within the report and $N_R$ is the length of the report. We formulate our approach as:
\begin{gather}
K_c = \mathrm{WCK}(C), \\
K_t = \mathrm{MRK}(I, D), \\
R = \mathrm{Transformer}(\mathrm{MoK}(F_I, K_c, K_t)).
\end{gather}

\noindent Specifically, $\mathrm{WCK}(\cdot)$ represents the Weighted Concept Knowledge branch, where $C$ denotes the predefined clinical concepts package. $K_c$ stands for the weighted and class-balanced clinical knowledge. $\mathrm{MRK}(\cdot)$ signifies the Multimodal Retrieval Knowledge branch, with $D$ representing the database, and $K_t$ denoting the set of triplets corresponding to reports retrieved for the current image. $\mathrm{MoK}(\cdot)$ denotes the Mixture of Knowledge module, which combines the current image with the two knowledge sources before sending them to the Transformer for report generation.

Given the ground truth report $ R^*=\{ y^*_1, y^*_2, \ldots ,y^*_{N_R} \}$, we can train the model by minimizing the cross-entropy loss:

\begin{gather}
\mathcal{L}_{\mathrm{CE}}(\theta)= -\sum_{i=1}^{N_R} \log p_{\theta}(y_i=y^*_i|y^*_{1:i-1},I).
\end{gather}

In the following subsections, we will provide a detailed explanation of the extraction processes for these two types of knowledge, as well as the final fusion process.

\subsection{Weighted Concept Knowledge}
Unlike natural image descriptions, radiology reports contain rich specialized terms and clinical knowledge. Merely learning from the dataset is inadequate to equip the model with the ability to comprehend and apply these specialized terms. In contrast to the conventional approach~\cite{yang2023radiology} of using 14 sparse labels as external knowledge, we construct a more extensive collection of clinical concepts, comprising the 76 most frequently encountered lesions in radiology reports (e.g., normal, atelectasis, and consolidation). Enriched lesion category improves the descriptive precision of generated reports.

However, another critical challenge arises when dealing with clinical concepts. As illustrated in Fig.~\ref{fig:illustration_samples}, there is a  class imbalance problem in medical reports. If these 76 clinical concepts are considered equally important and fused with medical images, the model might still disproportionately focus on $normal$. To address this, we innovatively introduce the TF-IDF methodology~\cite{salton1988term} from the information retrieval domain to re-weight these concepts.

Specifically, for the current image, the weight score of each word in its corresponding report is first calculated as:
\begin{gather}
score_{i,j} = \frac{n_{i,j}}{\sum_{k}^{} n_{k,j}} \times \log \frac{|R|}{1+|j:w_i \in r_j|},
\end{gather}

\noindent where $n_{i,j}$ represents the frequency of word $w_{i}$ appearing in report $r_{j}$, $|R|$ denotes the total number of reports in the report corpus, and $|{j : w_i \in r_j}|$ is the count of reports containing $w_i$. 

During testing, as the ground truth report is not accessible, we calculate the weights using a straightforward retrieve-then-merge strategy. Specifically, we retrieve the Top-$k$ similar samples associated with the test image and compute the average TF-IDF weights of the corresponding retrieved images.

Given the clinical concept package $C = [c_1, c_2, ..., c_{N_c}]$, where $N_c$ denotes the total number of clinical concepts, we use ClinicalBert~\cite{alsentzer2019publicly} to obtain the clinical concept feature $F_c \in \mathbb{R}^{B \times N_c \times d}$, where $B$ and $d$ denote the batch size and dimension of the feature, respectively.
The corresponding weights of $C$ are denoted as $S_c=[s_1, s_2,..., s_{N_c}]$. The calculation manner for each $s_k$ is formulated as: 

\begin{equation}
s_k = \left\{
\begin{aligned}
    \begin{array}{@{}c@{}l}
        score_{g(k), j} &, \  \mathrm{if} \  c_k \in r_j , \\
        0 &, \  \mathrm{otherwise}. \\
    \end{array}
\end{aligned}
\right.
\end{equation}

\noindent where $g(k)$ is a mapping function which indicates the index of concept $c_k$ in the original report.

After obtaining the weight scores $S_c\in \mathbb{R}^{B \times N_c \times 1}$, we expand their dimensions to $d$, and multiply them with the clinical concept features $F_c$ to obtain the final weighted concept knowledge $K_c$. 
\begin{equation}
K_c = F_c \odot S_c.
\end{equation}

The TF-IDF score of each concept increases proportionally with their frequency in the current report, but decreases inversely with their frequency across the entire corpus. This approach allows us to identify the corresponding concepts while also addressing the issue of data imbalance.

\subsection{Multimodal Retrieval Knowledge}
\label{sec:multimodal_retrieval_knowledge}

Due to the standardized and highly overlapping nature of medical image descriptions compared to natural images, we can effectively leverage report information from other medical images to enhance the generation of reports for the current image. To accomplish this retrieval task, we carefully select MGCA~\cite{wang2022multi} as our frozen pre-trained vision-language model. Firstly, we extract the features of medical images in database $D$ and calculate their cosine similarity to determine the Top-$k$ samples that are most similar to the current image.

\begin{gather}
F_R = \mathrm{MGCA}(I),\ s(I_1, I_2)=\cos(F_{R_1}, F_{R_2})
\end{gather}

After retrieving the Top-$k$ most similar images to the current image, we extract their corresponding reports. Based on the retrieved reports extracted above, we introduce a novel triplet format \{entity, position, exist\} for the first time in the radiology report generation task. In addition to highlighting entities, such triplet format emphasizes their specific positions and existence. Compared to direct utilization of raw reports, this strategy avoids unnecessary complexity stemming from linguistic grammar, thus enhancing the supervision signals. Thus we can obtain more concise and clear knowledge that can be easily assimilated and utilized by subsequent parts of the network.

\begin{figure}[h]
\centering
\includegraphics[width=0.46\textwidth]{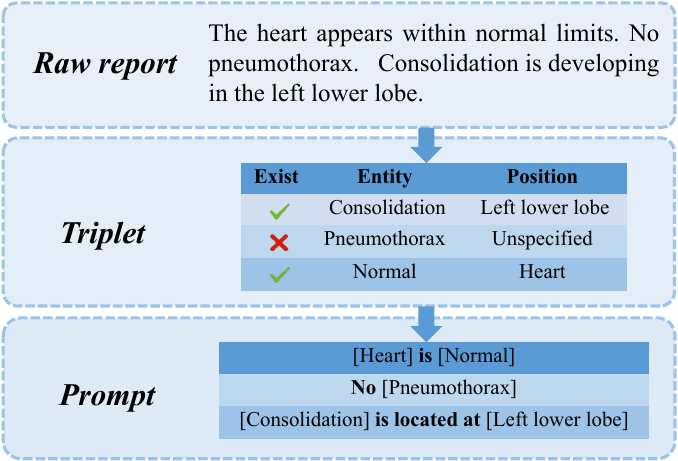}
\caption{An illustration of the triplet extraction process.}
\label{fig:triplet_extraction}
\vspace{-0.4cm}
\end{figure}

The complete process is illustrated in Fig.~\ref{fig:triplet_extraction}, where we utilize Rad-Graph~\cite{jain2021radgraph} and AGXNet~\cite{yu2022anatomy} as our tools for extracting triplets. 
For any retrieved report, we first extract medical keywords from the sentences and categorize them into [entity] or [position]. Here, [entity] represents clinical observations such as $normal$ and $enlarged$, while [position] denotes the anatomical body part referenced in a radiology report, like $right \ lower \ lobe$. 
This anatomical localization is crucial for clinical lesion identification.

\vspace{5mm}
Next, we annotate the entities as ``exist'' and ``absent'' to represent the existence information in the triplet. Subsequently, these entities are combined with their corresponding positions and existence states to form triplets. Lastly, to facilitate the effective encoding and fusion of these triplets, we establish a well-designed prompt mechanism. This mechanism provides different prompt strategies based on the distinct grammatical properties of entities (such as $noun$ or $adjective$):

\begin{equation}
\left\{
\begin{aligned}
    & \mathrm{No} \ [\mathrm{entity}], \\
    & [\mathrm{position}]\ \mathrm{is} \ [\mathrm{entity}], \\
    & [\mathrm{entity}] \ \mathrm{is} \ \mathrm{located} \ \mathrm{at} \ [\mathrm{position}].
\end{aligned}
\right.
\end{equation}

\begin{figure*}[h]
\centering
\includegraphics[width=0.95\textwidth]{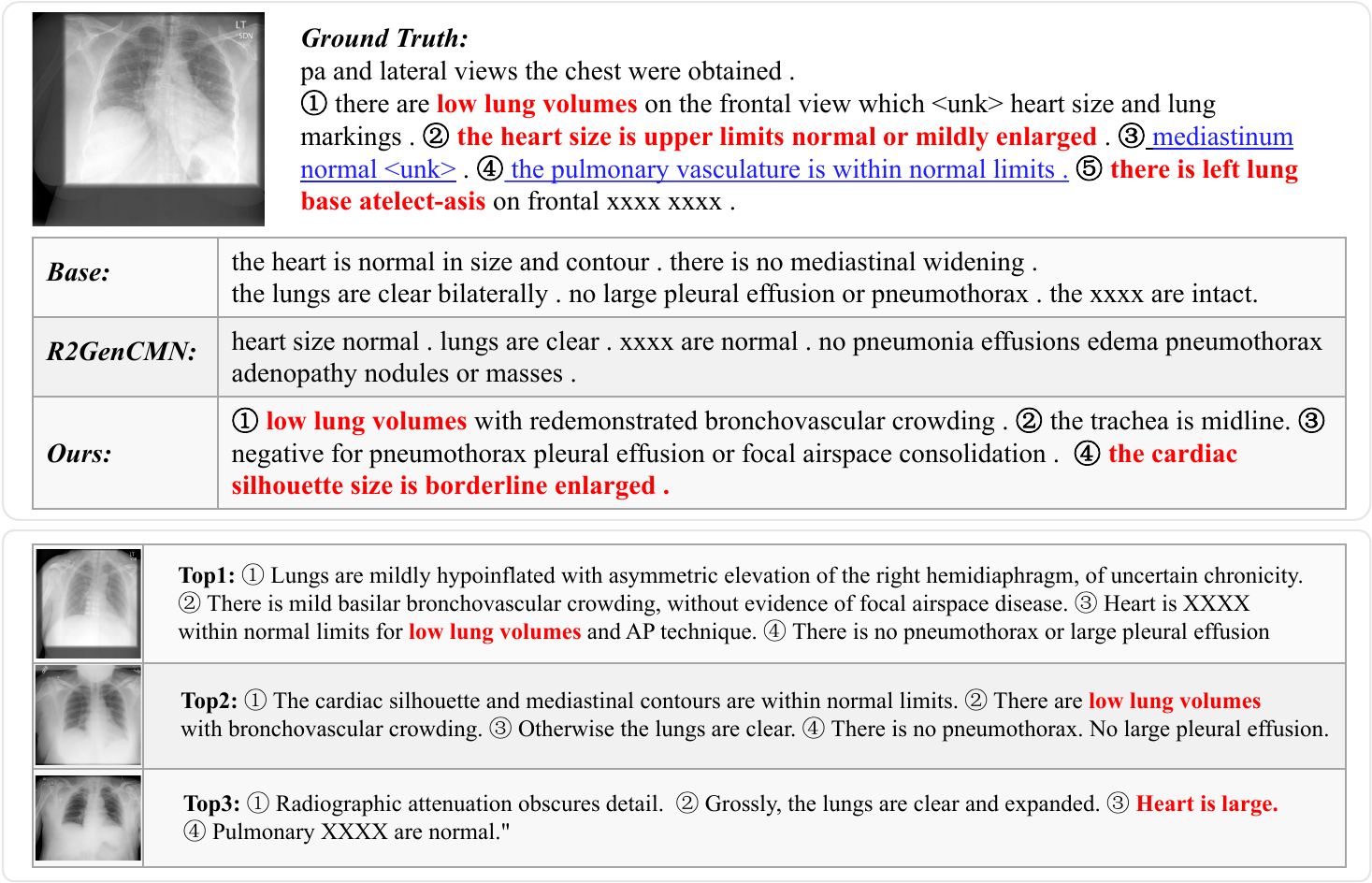}
\caption{Reports generated by different methods and the retrieval results of our method. In report retrieval, we highlight both \textcolor{blue}{\underline{normal}} and \textcolor{red}{\textbf{abnormal}}.}
\label{fig:visualization}
\end{figure*}

The Multimodal Retrieval Knowledge branch can avoid the unnecessary complexity of grammatical understanding, while still retaining the valuable information in reports. Thus the maximum extraction of valuable, accurate, and non-redundant knowledge can be ensured.

\subsection{Mixture of Knowledge and Report Generation}
By employing the aforementioned two knowledge branches, we acquire fine-grained and precise domain-specific knowledge. Subsequently, the MoK module is exploited to embed two types of knowledge. It takes the image feature as query, and cross attends to weighted concept knowledge feature ($K_c$) and multimodal retrieval knowledge feature ($K_t$), which can be formulated as:
\begin{gather}
F_{I'} = F_I + \mathrm{Att}(F_I,K_c,K_c) +\mathrm{ Att}(F_I,K_t,K_t), \\
\mathrm{Att}(Q,K,V) = \mathrm{Softmax}(QK^\top/\sqrt{d})V.
\end{gather}
Once the weighted image feature is obtained, it serves as input to the encoder, and is then fed into the Transformer decoder to produce the final report.

\begin{table}[t]
\caption{Performance of the proposed method compared with existing methods on the test sets of IU-Xray and MIMIC-CXR.}
\label{tab:performance_comparison}
\centering
\resizebox{\linewidth}{!}{
\begin{tabular}{M{1pt}M{48pt}|ccccccc}
\hline
 & Method & BL-1 & BL-2 & BL-3 & BL-4 & RG-L & MTOR & CIDEr \\
\hline
\multirow{10}{*}{\rotatebox{90}{IU-Xray}} & CoAttn~\cite{jing2018automatic} & 0.455 & 0.288 & 0.205 & 0.154 & 0.369 & - & 0.277 \\
 & KERP~\cite{li2019knowledge} & 0.482 & 0.325 & 0.226 & 0.162 & 0.339 & - & 0.280 \\
 & R2Gen~\cite{chen2020generating} & 0.470 & 0.304 & 0.219 & 0.165 & 0.371 & 0.187 & - \\
 & PPKED~\cite{liu2021exploring} & 0.483 & 0.315 & 0.224 & 0.168 & 0.376 & 0.190 & 0.351 \\
 & R2GenCMN~\cite{chen2021cross} & 0.470 & 0.304 & 0.222 & 0.179 & 0.358 & 0.191 & 0.344 \\
 & Know Mat~\cite{yang2022knowledge} & 0.496 & 0.327 & 0.238 & 0.178 & 0.381 & - & - \\
 & CMM+RL~\cite{qin2022reinforced} & 0.494 & 0.321 & 0.235 & 0.181 & 0.384 & 0.201 & - \\
 & M2KT~\cite{yang2023radiology} & 0.497 & 0.319 & 0.230 & 0.174 & 0.399 & - & 0.407 \\ 
 & DCL~\cite{li2023dynamic} & - & - & - & 0.163 & 0.383 & 0.193 & 0.586 \\
 & Ours & \textbf{0.516} & \textbf{0.349} & \textbf{0.262} & \textbf{0.207} & \textbf{0.400} & \textbf{0.222} & \textbf{0.608} \\
\hline
\multirow{9}{*}{\rotatebox{90}{MIMIC-CXR}} & TopDown~\cite{anderson2018bottom} & 0.317 & 0.195 & 0.130 & 0.092 & 0.267 & 0.128 & - \\
 & R2Gen~\cite{chen2020generating} & 0.353 & 0.218 & 0.145 & 0.103 & 0.277 & 0.142 & - \\
 & PPKED~\cite{liu2021exploring} & 0.360 & 0.224 & 0.149 & 0.106 & 0.284 & 0.149 & 0.237 \\
 & R2GenCMN~\cite{chen2021cross} & 0.353 & 0.218 & 0.148 & 0.106 & 0.278 & 0.142 & - \\
 & Know Mat~\cite{yang2022knowledge} & 0.363 & 0.228 & 0.156 & 0.115 & 0.284 & - & 0.203 \\
 & CMM+RL~\cite{qin2022reinforced} & 0.381 & 0.232 & 0.155 & 0.109 & 0.287 & 0.151 & - \\
 & M2KT~\cite{yang2023radiology} & \textbf{0.386} & \textbf{0.237} & 0.157 & 0.111 & 0.274 & - & 0.111 \\
 & DCL~\cite{li2023dynamic} & - & - & - & 0.109 & 0.284 & 0.150 & \textbf{0.281} \\
 & Ours & 0.360 & 0.231 & \textbf{0.162} & \textbf{0.119} & \textbf{0.298} & \textbf{0.153} & 0.217 \\
\hline
\end{tabular}}
\end{table}

\section{Experiments}
\subsection{Datasets}
We conduct experiments on two public benchmarks, namely IU-Xray~\cite{demner2016preparing} and MIMIC-CXR~\cite{johnson2019mimic}. To ensure a fair comparison, we adopt the settings in \cite{chen2021cross} for report preprocessing. 

\textbf{IU-Xray}, provided by Indiana University, is the most widely used public benchmark dataset for evaluating the performance of radiology report generation methods. It contains 3,955 reports, each paired with two X-ray images, resulting in a total of 7,470 images.   Following \cite{chen2021cross}, we split the dataset into train/validation/test sets with a ratio of 7:1:2. 

\textbf{MIMIC-CXR}, provided by the Beth Israel Deaconess Medical Center, is the largest public dataset in report generation. It contains 377,110 chest X-ray images and 227,835 corresponding reports from 64,588 patients. We follow the official train/validation/test splits in~\cite{johnson2019mimic}.

\subsection{Metrics and Settings}
\textbf{Metrics.} To evaluate model performance, we employ widely-used Natural Language Generation (NLG) metrics: BLEU (BL)~\cite{papineni2002bleu}, METEOR (MTOR)~\cite{banerjee2005meteor}, ROUGE-L (RG-L)~\cite{rouge2004package} and CIDEr~\cite{vedantam2015cider}.

\textbf{Settings.} In our implementation, all images are resized to 224 $\times$ 224. Following previous work~\cite{chen2021cross}, we use frontal and lateral X-ray images as input for IU-Xray, and only frontal X-ray images for MIMIC-CXR. For the visual extractor, we adopt the ResNet101~\cite{he2016deep} pretrained on ImageNet. As for the encoder-decoder backbone, we use a randomly initialized Transformer with 3 layers. Knowledge embedding uses the pre-trained ClinicalBERT~\cite{alsentzer2019publicly} model. For retrieving similar images, we use the MGCA~\cite{wang2022multi} pre-trained on MIMIC-CXR dataset, and set the Top-$k$ to 3. The training process involves 100 epochs for IU-Xray and 30 epochs for MIMIC-CXR. We adopt Adam optimizer with an initial learning rate of $1 \times 10^{-4}$ and a weight decay of $5 \times 10^{-5}$ for training, except for the visual extractor which has an initial learning rate of $5 \times 10^{-5}$.

\subsection{Comparison with State-of-the-Art Methods}
\textbf{Quantitative Analysis.} To demonstrate its effectiveness, we compare our method with a wide range of SOTA methods on both IU-Xray and MIMIC-CXR. As shown in Table~\ref{tab:performance_comparison}, our approach achieves competitive performance on both datasets. Especially, on IU-Xray dataset, BL-3 and BL-4 have increased by 2.4\% and 2.6\% respectively; on MIMIC-CXR dataset, they have increased by 0.5\% and 0.4\%, this demonstrates that our model can generate more accurate and fluent reports. Compared to IU-Xray, MIMIC-CXR's reports are longer and include more enumerations of normal phenomena. However, the determination of describing normal regions depends on radiologists and is arbitrary, as reports do not require mentioning non-pathological areas. Therefore, we believe that this rather subjective decision-making cannot be learned by the model, which may result in relatively lower BL-1 and CIDEr scores.

\textbf{Qualitative Analysis.} We select a challenging case from the test set for illustration. The first block of Fig.~\ref{fig:visualization} shows the ground truth of the target image, with subscripts denoting sentences containing entity descriptions. Below the image are the report generation results of the Baseline Model (referring to using only the Transformer encoder-decoder without knowledge injection), R2GenCMN and our method. It can be observed that the first two methods do not focus on the abnormal regions in the image. In contrast, our generated report contains both \textit{low lung volumes} and \textit{the cardiac silhouette size is borderline enlarged}, and it is more fluent, without nonsense information such as \textit{xxxx}. This demonstrates that our method has the capability to accurately identify and describe lesions.

In the second block of Fig.~\ref{fig:visualization}, the model's Top-3 retrieved similar reports are presented. It can be seen that these reports capture certain aspects of the lesions featured in the target image, providing valuable guidance for the final report generation. This underscores the significance of incorporating retrieval-based triplets in our approach.

\subsection{Ablation Study}

\begin{table}[h]
\caption{Ablation study of our method on IU-Xray dataset.}
\label{tab:whole_model}
\footnotesize
\resizebox{\columnwidth}{!}{
\begin{tabular}{cccccccc}
\hline
Model & BL-1 & BL-2 & BL-3 & BL-4 & MTOR  & RG-L   & CIDEr \\ \hline
\textbf{Base}                 & 0.466 & 0.300 & 0.219 & 0.168 & 0.188 & 0.360 & 0.427 \\
\textbf{+Concepts}            & 0.477 & 0.314 & 0.229 & 0.174 & 0.200 & 0.376 & 0.527 \\
\textbf{+We\_Conp}        & 0.492 & 0.326 & 0.241 & 0.189 & 0.207 & 0.380 & \textbf{0.629} \\
\textbf{+Triplet}         & 0.500 & 0.323 & 0.231 & 0.171 & 0.211 & 0.379 & 0.524 \\
\textbf{Ours}                & \textbf{0.516} & \textbf{0.349} & \textbf{0.262} & \textbf{0.207} & \textbf{0.222} & \textbf{0.400} & 0.608 \\ \hline
\end{tabular}
}
\end{table}

We conduct ablation experiments on IU-Xray to validate the efficacy of our architecture. As shown in Table~\ref{tab:whole_model}, we start from a basic encoder-decoder framework, and then inject concepts, weighted concepts and triples, respectively. It can be observed that adding concepts and triplets individually, in comparison to the baseline model, leads to significant improvements on all metrics. This observation indicates the effectiveness of both forms of knowledge injection. Triplet-based augmentation exhibits greater enhancement compared to single conceptual entities. This might stem from the fact that triplets, obtained through retrieval, are more closely related to the current image. These triplets contain additional detailed information, such as position and existence. Incorporating weighted concepts shows substantial growth, with the introduction of TF-IDF scores effectively mitigating the issue of class imbalance. This encourages the model to focus more on abnormal regions and images with anomalies. Ultimately, the integration of such three improvements yields the best overall performance.

\section{Conclusion}
In this paper, we propose an enhanced knowledge injection framework for radiology report generation. Based on the weighted concept knowledge and multimodal retrieval knowledge, we mitigate the problem of class imbalance while injecting well-structured and low-noise knowledge. Integrating these two types of knowledge with the current image allows us to leverage their respective strengths, facilitating accurate radiology report generation. Our future work will mainly focus on improving the accuracy of retrieval and exploring other effective knowledge sources.

\bibliographystyle{IEEEtran}
\bibliography{IEEEabrv,mybib}

\begin{thebibliography}{10}
\providecommand{\url}[1]{#1}
\csname url@samestyle\endcsname
\providecommand{\newblock}{\relax}
\providecommand{\bibinfo}[2]{#2}
\providecommand{\BIBentrySTDinterwordspacing}{\spaceskip=0pt\relax}
\providecommand{\BIBentryALTinterwordstretchfactor}{4}
\providecommand{\BIBentryALTinterwordspacing}{\spaceskip=\fontdimen2\font plus
\BIBentryALTinterwordstretchfactor\fontdimen3\font minus \fontdimen4\font\relax}
\providecommand{\BIBforeignlanguage}[2]{{%
\expandafter\ifx\csname l@#1\endcsname\relax
\typeout{** WARNING: IEEEtran.bst: No hyphenation pattern has been}%
\typeout{** loaded for the language `#1'. Using the pattern for}%
\typeout{** the default language instead.}%
\else
\language=\csname l@#1\endcsname
\fi
#2}}
\providecommand{\BIBdecl}{\relax}
\BIBdecl

\bibitem{wang2018tienet}
X.~Wang, Y.~Peng, L.~Lu, Z.~Lu, and R.~M. Summers, ``Tienet: Text-image embedding network for common thorax disease classification and reporting in chest x-rays,'' in \emph{Proceedings of the IEEE Conference on Computer Vision and Pattern Recognition}, 2018, pp. 9049--9058.

\bibitem{jing2018automatic}
B.~Jing, P.~Xie, and E.~Xing, ``On the automatic generation of medical imaging reports,'' in \emph{Proceedings of the 56th Annual Meeting of the Association for Computational Linguistics (Volume 1: Long Papers)}, 2018, pp. 2577--2586.

\bibitem{li2018hybrid}
Y.~Li, X.~Liang, Z.~Hu, and E.~P. Xing, ``Hybrid retrieval-generation reinforced agent for medical image report generation,'' \emph{Advances in Neural Information Processing Systems}, vol.~31, 2018.

\bibitem{liu2021exploring}
F.~Liu, X.~Wu, S.~Ge, W.~Fan, and Y.~Zou, ``Exploring and distilling posterior and prior knowledge for radiology report generation,'' in \emph{Proceedings of the IEEE/CVF Conference on Computer Vision and Pattern Recognition}, 2021, pp. 13\,753--13\,762.

\bibitem{chen2021cross}
Z.~Chen, Y.~Shen, Y.~Song, and X.~Wan, ``Cross-modal memory networks for radiology report generation,'' in \emph{Proceedings of the 59th Annual Meeting of the Association for Computational Linguistics and the 11th International Joint Conference on Natural Language Processing (Volume 1: Long Papers)}, 2021, pp. 5904--5914.

\bibitem{qin2022reinforced}
H.~Qin and Y.~Song, ``Reinforced cross-modal alignment for radiology report generation,'' in \emph{Findings of the Association for Computational Linguistics: ACL 2022}, 2022, pp. 448--458.

\bibitem{yang2022knowledge}
S.~Yang, X.~Wu, S.~Ge, S.~K. Zhou, and L.~Xiao, ``Knowledge matters: Chest radiology report generation with general and specific knowledge,'' \emph{Medical Image Analysis}, vol.~80, p. 102510, 2022.

\bibitem{li2023dynamic}
M.~Li, B.~Lin, Z.~Chen, H.~Lin, X.~Liang, and X.~Chang, ``Dynamic graph enhanced contrastive learning for chest x-ray report generation,'' in \emph{Proceedings of the IEEE/CVF Conference on Computer Vision and Pattern Recognition}, 2023, pp. 3334--3343.

\bibitem{li2019knowledge}
C.~Y. Li, X.~Liang, Z.~Hu, and E.~P. Xing, ``Knowledge-driven encode, retrieve, paraphrase for medical image report generation,'' in \emph{Proceedings of the AAAI Conference on Artificial Intelligence}, vol.~33, no.~01, 2019, pp. 6666--6673.

\bibitem{zhang2020radiology}
Y.~Zhang, X.~Wang, Z.~Xu, Q.~Yu, A.~Yuille, and D.~Xu, ``When radiology report generation meets knowledge graph,'' in \emph{Proceedings of the AAAI Conference on Artificial Intelligence}, vol.~34, no.~07, 2020, pp. 12\,910--12\,917.

\bibitem{yang2023radiology}
S.~Yang, X.~Wu, S.~Ge, Z.~Zheng, S.~K. Zhou, and L.~Xiao, ``Radiology report generation with a learned knowledge base and multi-modal alignment,'' \emph{Medical Image Analysis}, vol.~86, p. 102798, 2023.

\bibitem{salton1988term}
G.~Salton and C.~Buckley, ``Term-weighting approaches in automatic text retrieval,'' \emph{Information Processing \& Management}, vol.~24, no.~5, pp. 513--523, 1988.

\bibitem{alsentzer2019publicly}
E.~Alsentzer, J.~Murphy, W.~Boag, W.-H. Weng, D.~Jindi, T.~Naumann, and M.~McDermott, ``Publicly available clinical bert embeddings,'' in \emph{Proceedings of the 2nd Clinical Natural Language Processing Workshop}, 2019, pp. 72--78.

\bibitem{wang2022multi}
F.~Wang, Y.~Zhou, S.~Wang, V.~Vardhanabhuti, and L.~Yu, ``Multi-granularity cross-modal alignment for generalized medical visual representation learning,'' \emph{Advances in Neural Information Processing Systems}, vol.~35, pp. 33\,536--33\,549, 2022.

\bibitem{jain2021radgraph}
S.~Jain, A.~Agrawal, A.~Saporta, S.~Truong, T.~Bui, P.~Chambon, Y.~Zhang, M.~P. Lungren, A.~Y. Ng, C.~Langlotz \emph{et~al.}, ``Radgraph: Extracting clinical entities and relations from radiology reports,'' in \emph{Thirty-fifth Conference on Neural Information Processing Systems Datasets and Benchmarks Track (Round 1)}, 2021.

\bibitem{yu2022anatomy}
K.~Yu, S.~Ghosh, Z.~Liu, C.~Deible, and K.~Batmanghelich, ``Anatomy-guided weakly-supervised abnormality localization in chest x-rays,'' in \emph{International Conference on Medical Image Computing and Computer-Assisted Intervention}.\hskip 1em plus 0.5em minus 0.4em\relax Springer, 2022, pp. 658--668.

\bibitem{chen2020generating}
Z.~Chen, Y.~Song, T.-H. Chang, and X.~Wan, ``Generating radiology reports via memory-driven transformer,'' in \emph{Proceedings of the 2020 Conference on Empirical Methods in Natural Language Processing (EMNLP)}, 2020, pp. 1439--1449.

\bibitem{anderson2018bottom}
P.~Anderson, X.~He, C.~Buehler, D.~Teney, M.~Johnson, S.~Gould, and L.~Zhang, ``Bottom-up and top-down attention for image captioning and visual question answering,'' in \emph{Proceedings of the IEEE Conference on Computer Vision and Pattern Recognition}, 2018, pp. 6077--6086.

\bibitem{demner2016preparing}
D.~Demner-Fushman, M.~D. Kohli, M.~B. Rosenman, S.~E. Shooshan, L.~Rodriguez, S.~Antani, G.~R. Thoma, and C.~J. McDonald, ``Preparing a collection of radiology examinations for distribution and retrieval,'' \emph{Journal of the American Medical Informatics Association}, vol.~23, no.~2, pp. 304--310, 2016.

\bibitem{johnson2019mimic}
A.~E. Johnson, T.~J. Pollard, N.~R. Greenbaum, M.~P. Lungren, C.-y. Deng, Y.~Peng, Z.~Lu, R.~G. Mark, S.~J. Berkowitz, and S.~Horng, ``Mimic-cxr-jpg, a large publicly available database of labeled chest radiographs,'' \emph{arXiv preprint arXiv:1901.07042}, 2019.

\bibitem{papineni2002bleu}
K.~Papineni, S.~Roukos, T.~Ward, and W.-J. Zhu, ``Bleu: a method for automatic evaluation of machine translation,'' in \emph{Proceedings of the 40th Annual Meeting of the Association for Computational Linguistics}, 2002, pp. 311--318.

\bibitem{banerjee2005meteor}
S.~Banerjee and A.~Lavie, ``Meteor: An automatic metric for mt evaluation with improved correlation with human judgments,'' in \emph{Proceedings of the ACL Workshop on Intrinsic and Extrinsic Evaluation Measures for Machine Translation and/or Summarization}, 2005, pp. 65--72.

\bibitem{rouge2004package}
L.~C. ROUGE, ``A package for automatic evaluation of summaries,'' in \emph{Proceedings of Workshop on Text Summarization of ACL, Spain}, vol.~5, 2004.

\bibitem{vedantam2015cider}
R.~Vedantam, C.~Lawrence~Zitnick, and D.~Parikh, ``Cider: Consensus-based image description evaluation,'' in \emph{Proceedings of the IEEE Conference on Computer Vision and Pattern Recognition}, 2015, pp. 4566--4575.

\bibitem{he2016deep}
K.~He, X.~Zhang, S.~Ren, and J.~Sun, ``Deep residual learning for image recognition,'' in \emph{Proceedings of the IEEE Conference on Computer Vision and Pattern Recognition}, 2016, pp. 770--778.

\end{thebibliography}

\end{document}